\documentclass[10pt,journal,compsoc]{IEEEtran}

\usepackage{times}
\usepackage{epsfig}

\usepackage[colorlinks]{hyperref}

\usepackage{nicefrac}       
\usepackage{amsthm,amssymb,amsmath,bm}
\usepackage{xcolor}        
\usepackage{subcaption,graphicx}
\usepackage{listings}
\usepackage{multirow}
\usepackage{algorithmic}
\usepackage{algorithm}
\usepackage{diagbox}
\usepackage{colortbl}
\definecolor{codeAnnotation}{RGB}{106,153,85}
\definecolor{rowBackgroundColor}{RGB}{242,242,255}

\allowdisplaybreaks[4]
\usepackage{pifont}
\usepackage{nicematrix}

\usepackage{bbding}

\usepackage[compress]{cite}

\usepackage{ragged2e}

\usepackage[utf8]{inputenc} 
\usepackage[T1]{fontenc}    
\usepackage{hyperref}       
\usepackage{url}            
\usepackage{booktabs}       
\usepackage{amsfonts}       
\usepackage{nicefrac}       
\usepackage{microtype}      
\usepackage{xcolor}         

\usepackage{caption}
\usepackage{wrapfig}
\usepackage{amssymb,amsmath,bm}
\usepackage{subcaption,graphicx}
\usepackage{multirow}
\usepackage{bbm}
\usepackage{algorithmic}
\usepackage{algorithm}
\usepackage{diagbox}
\definecolor{codeAnnotation}{RGB}{106,153,85}

\allowdisplaybreaks[4]

\begin{document}

\title{Variational Positive-incentive Noise: How Noise Benefits Models} 

\author{
Hongyuan Zhang, 
Sida Huang, 
Yubin Guo,
and Xuelong Li$^*$, \IEEEmembership{~Fellow,~IEEE} 
\thanks{
$^*$: Corresponding author. 
}

\thanks{
The authors are with the Institute of Artificial Intelligence (TeleAI), China Telecom, P. R. China. 
Hongyuan Zhang is also with The University of Hong Kong. 
Sida Huang is also with the School of Artificial Intelligence, OPtics and ElectroNics (iOPEN), Northwestern Polytechnical University, Xi'an 710072, P.R. China. 
}


\thanks{E-mail: hyzhang98@gmail.com, sidahuang2001@gmail.com, g2247157972@gmail.com, xuelong\_li@ieee.org.}

}

\markboth{IEEE TRANSACTIONS ON PATTERN ANALYSIS AND MACHINE INTELLIGENCE}{Zhang \MakeLowercase{\textit{et al.}}: 
Variational Positive-incentive Noise: How Noise Benefits Models
}

\IEEEtitleabstractindextext{
\justifying  

\begin{abstract}
  A large number of works aim to alleviate the impact of noise 
  due to an underlying conventional assumption of the negative role of noise. 
  However, some existing works show that the assumption does not always hold. 
  In this paper, we investigate how to benefit the classical models by random noise 
  under the framework of Positive-incentive Noise (\textit{Pi-Noise}) \cite{Pi-Noise}. 
  Since the ideal objective of Pi-Noise is intractable, we propose to optimize its variational bound instead, 
  namely variational Pi-Noise (\textit{VPN}). 
  With the variational inference, a VPN generator implemented by neural networks is designed for 
  enhancing base models and simplifying the inference of base models, 
  without changing the architecture of base models. 
  Benefiting from the independent design of base models and VPN generators, 
  the VPN generator can work with most existing models. 
  From the extensive experiments on different base models (including linear models, ResNet, ViT, \textit{etc.}) 
  it is shown that the proposed VPN generator can improve the base models. 
  It is appealing that the trained VPN generator prefers 
  to blur the irrelevant ingredients in complicated images, 
  which meets our expectations. 
\end{abstract}

\begin{IEEEkeywords}
  Positive-incentive Noise, Beneficial Noise, Variational Inference.
\end{IEEEkeywords}

}

\maketitle

\section{Introduction} \label{section_intro}
Although there are plenty of works aiming to enhance the deep models via 
eliminating the noise contained in data, some studies \cite{Dropout,AdversarialTraining,GAN,NCE,DDPM,LabelSmoothing} have explicitly or implicitly 
shown the potential positive effect of noise in plenty of topics, 
including but not limited to video diffusion \cite{DiffusionNoise}, image-to-image translation \cite{ImageTranslationNoise}, 
and point cloud segmentation \cite{PointCloudNoise}. 
Different from most works that empirically and heuristically utilize noise, the Positive-incentive Noise (\textit{Pi-Noise} or \textit{$\pi$-noise}) \cite{Pi-Noise} is 
a new framework using information theory that scientifically investigates the positive or negative impact brought by noise. 
Motivated by $\pi$-noise, 
we focus on \textbf{how to apply the beneficial noise to deep learning models} with $\pi$-noise framework , 
which formally analyzes noise using information theory. 
The noise that simplifies the task is defined as $\pi$-noise. 
Formally speaking, the noise $\bm \varepsilon \in \mathcal{E}$ 
(where $\mathcal{E}$ is a noise set) 
satisfying
\begin{equation}
  I(\mathcal T, \mathcal{E}) > 0 \Leftrightarrow H(\mathcal{T}) > H(\mathcal{T} | \mathcal{E}) 
\end{equation}
is defined as $\pi$-noise to the task $\mathcal{T}$, 
where $H(\cdot)$ represents the information entropy and 
$I(\cdot, \cdot)$ denotes the mutual information. 
Noise is named as pure noise provided that $I(\mathcal{T}, \mathcal{E})=0$. 
Although adding some random noise is sometimes regarded as an optional data augmentation method, 
it is an unstable scheme in practice. 
Following this principle, 
it becomes possible to definitely predict whether a class of distributions 
would benefit the target task $\mathcal{T}$.
It is counterintuitive and attractive that some random noise would benefit the task, 
rather than disturbing it. 

The crux of $\pi$-noise is how to properly define the task entropy 
$H(\mathcal{T})$ and efficiently calculate it. 
As a basic task of machine learning, the task of single-label classification is relatively easy and it is direct to model its task entropy. 
It may be a rational scheme to measure its complexity 
by computing the uncertainty of labels. In other words, the entropy of $p(y | \bm x)$ 
is a crucial quantity related to the task complexity and it is formulated as 
\begin{equation} \label{eq_initial_definition_of_task_entropy}
  H(p(y|\bm x)) = \int -p(y | \bm x) \log p(y | \bm x) d y . 
\end{equation}
Remark that $y$ is assumed as a continuous variable for mathematical convenience, which implies the infinite classes, 
and the integral operation can be simply substituted by the summation 
if there are finite number of classes. 
In most scenarios, the labels annotated by humans are regarded 
as the true $p(y | \bm x)$. In other words, $p(y | \bm x)$ is just a one-hot vector. 
On more complicated datasets such as ImageNet \cite{ImageNet}, 
the one-hot property may be not the optimal setting for $p( y | \bm x)$. 
For example, an image with complicated background may be assigned 
to diverse labels with different probabilities. 
It should be pointed out that $p(y | \bm x)$ is different from multi-label classification \cite{MultiLabel}, 
which allows assigning multiple labels to a data point. 
This paper focuses on the single-label classification and the purpose of defining $p(y | \bm x)$ 
is to better predict a label for each sample. 

\begin{figure*}[t]
  \centering
  \includegraphics[width=\linewidth]{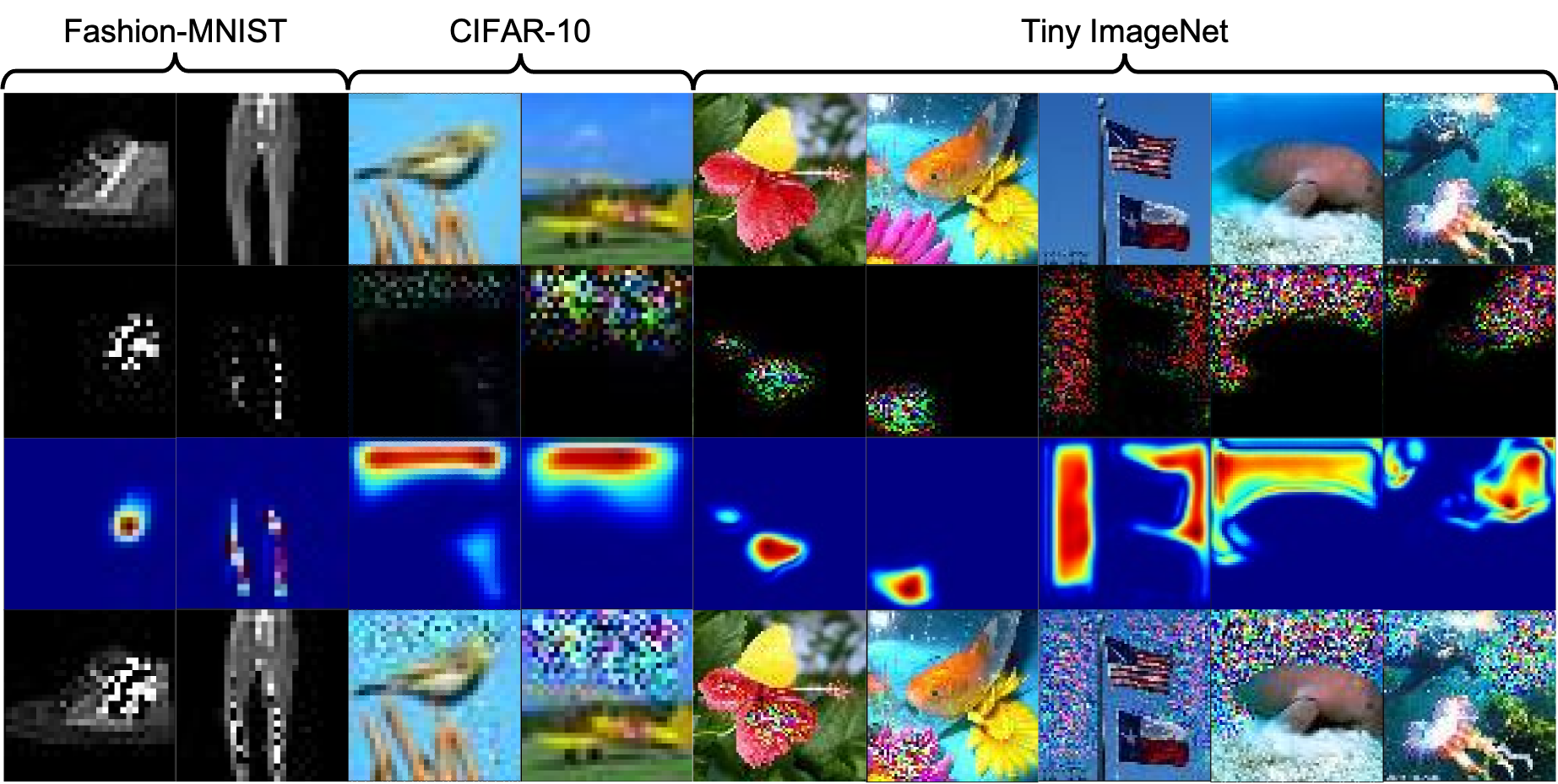}
  \caption{Visualization of generated $\pi$-noise. 
  The first line is the original image, 
  the second line shows the generated noise, 
  the third line is the heatmap of variance related to each pixel, 
  and the bottom line is the image with $\pi$-noise. 
  In the fifth image labeled by butterfly, the flower which is most similar to butterfly is 
  disturbed by intense noise so that the recognition task is significantly simplified. 
  }
  \label{figure_visualization}
\end{figure*}

The primary challenges of the $\pi$-noise principle are concluded as the following two aspects: 
\begin{itemize}
  \item [\textbf{\textit{C1}}:] Since there are integrals of several continuous variables (\textit{e.g.}, $\bm \varepsilon$) in mutual information, the computation is usually intractable. 
  \item [\textbf{\textit{C2}}:] Precise $p(y | \bm x)$ is unavailable since every data point has only one label as its ground truth on single-label classification datasets. 
\end{itemize}

In this paper, we apply the variational inference technique to $\pi$-noise, 
namely Variational Pi-Noise (\textit{VPN}). 
Via optimizing the variational bound, challenge \textit{C1} is addressed. 
Using the Monte Carlo method, 
we can train a $\pi$-noise generator without precise $p(y | \bm x)$, which partially addresses challenge \textit{C2}. 
The VPN generator learns the probability density function of noise 
and is implemented by neural networks. 
It can be applied to any existing models designed for task $\mathcal{T}$ (namely base models). 
A trained VPN generator can produce any amount of noise for any sample. 
It can be, thereby, used in both the training and inference phases of base models. 
The effectiveness of VPN is verified by experiments. 
From the experimental visualization shown by Figure \ref{figure_visualization}, we surprisingly find that the generated $\pi$-noise 
\textcolor{black}{
\textbf{prefers to learn to blur the irrelevant background of complicated images},
} 
which meets our expectation of the task entropy and \textcolor{black}{shows how the noise helps enhance the deep learning models}.

\textbf{Notations: } ~ 
In this paper, $\mathcal{X}$, $\mathcal{Y}$, and $\mathcal{E}$ represent 
the sample space, label space, and noise space, respectively. 
The vector is denoted by lower-case letter in bold and matrix is denoted 
by upper-case letter in bold. $\bm I$ is the identity matrix. 
$\mathcal{N}(\bm \mu, \bm \Sigma)$ is the multivariate Gaussian distribution 
with the mean vector $\bm \mu$ and correlation matrix $\bm \Sigma$. 
$x \rightarrow y \rightarrow z$ is a Markov chain. 
$\mathbb{R}_+$ is the set of non-negative real numbers. 
$\circ$ represents the composition operation of two functions and $\odot$ is the Hadamard product. 
We simply use $\int$ to represent both single and multiple Riemann integral, 
while $\int _E^{(L)} \cdot dx$ represents a Lebesgue integral on $E$.

\section{Related Work}

\subsection{Difference from Conventional Augmentation}
One may be confused about the difference between $\pi$-noise and conventional 
data augmentation \cite{AutoAugment} (such as translation, rotation, noising, and shearing of vision data), 
which are particularly prevalent on account of contrastive learning \cite{CPC,SimCLR,PiNDA}. 
The data augmentation only participates in the training phase 
and the testing sample is inputted without augmentation during the inference phase. 
In VPN, we train a generator that can produce $\pi$-noise for both the training and inference phases. 
In other words, one may use a well-trained VPN generator to automatically 
generate $\pi$-noise for any unseen samples and help the base model to distinguish 
the sophisticated samples, which is appealing compared with the conventional data augmentation. 

\begin{figure*}[t]
  \centering
  \includegraphics[width=\linewidth]{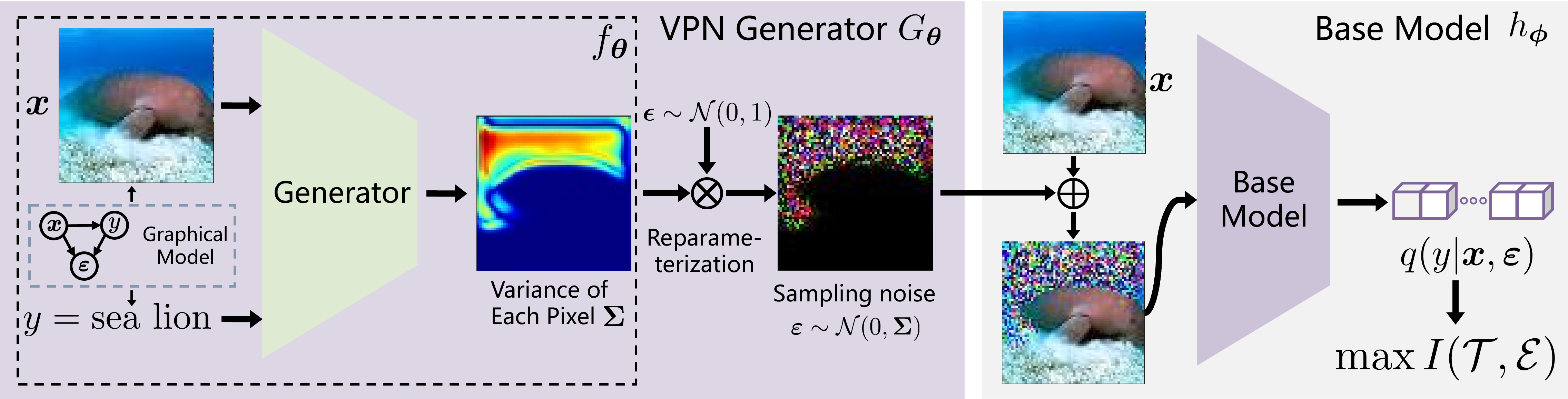}
  \caption{The illustration of VPN framework, which consists of a base model and a $\pi$-noise generator. 
  The generator can be trained either with the base model or after the base model. 
  Any model that can predict $p(y | \bm x)$ could be a valid base model. }
  \label{figure_framework}
  \vspace{-3mm}
\end{figure*}

\subsection{Learnable Augmentation}
Adversarial training \cite{AdversarialTraining} is another kind of popular learnable augmentation method \cite{VAT,IMSAT}, which is widely applied in recent years. 
It is a hot topic to enhance the robustness of deep models. 
It aims at adding perturbations (namely adversarial attack), during training deep models, to the input of neural network 
to lead the network to output the incorrect prediction with high confidence. 
For example, FSGM \cite{FSGM}, a white-box method, generates the attack regarding the gradient of $\bm x$. 
One pixel attack \cite{BlackBoxAttack}, a black-box method, only changes a pixel 
of an input image. 
The procedure of adversarial training is somewhat similar to $\pi$-noise. 
The primary difference is the purpose and impact of noise. 
The perturbations for adversarial attack are used to fool base models 
while $\pi$-noise is employed to enhance base models. 
From the optimization aspect, adversarial perturbations try to maximize the loss 
while $\pi$-noise can be regarded as minimizing it instead. 
Under the $\pi$-noise framework \cite{Pi-Noise}, adversarial perturbations should be classified as pure noise.

\subsection{Variational Inference}
The variational inference used in this paper is a practical technique in machine learning, 
especially in variational auto-encoder (VAE) \cite{VAE}, variational information bottleneck (VIB) \cite{VIB}, 
and their extensions \cite{GIB}. 
In particular, VIB attempts to maximize the mutual information between the label $y$
and latent representation $\bm z$ and meanwhile minimize the mutual information 
between $\bm z$ and input representation $\bm x$. 
However, VIB is essentially different from the proposed VPN. 
On the one hand, $y \rightarrow \bm x \rightarrow \bm z$ is a core assumption of VIB, 
while there is no assumption of conditional independence in the proposed VPN, 
which will be elaborated in the succeeding section. 
On the other hand, VIB essentially provides a new training principle for deep models 
to learn more informative and concise representations. 
In comparison, VPN introduces a new module, namely VPN generator. 
The architecture of VPN generator could be independent of the base deep models. 
The base models can either be co-trained with VPN generator or stay unchanged during training generator.

\section{Variational Positive-incentive Noise} \label{section_method}
It is our goal to design a practical paradigm for the generation of $\pi$-noise. 
And then the base models for task $\mathcal{T}$ can be fed with both the original data and learned $\pi$-noise. 
Remark that the added module learning $\pi$-noise would not change the architecture of base models. 
In summary, we expect to learn a probability density function (pdf) parameterized by $\bm \theta$, 
$\mathcal{D}_{\bm \theta}: \mathcal{X} \times \mathcal{Y} \mapsto \mathbb{R}_+$, 
which are trained with the base model, 
to produce $\pi$-noise that benefits the base model. 
Accordingly, the original training objective of learning $\pi$-noise is 
\begin{equation} \label{loss_original}
  \max_{\bm \theta} I(\mathcal{T}, \mathcal{E}), ~~ s.t. ~ \forall \bm x \in \mathcal{X}, y \in \mathcal{Y}, \bm \varepsilon \sim \mathcal{D}_{\bm \theta}(\bm x, y) . 
\end{equation}
The \textit{whole module}, consisting of the learning of $\mathcal{D}_{\bm \theta}$ and sampling from $\mathcal{D}_{\bm \theta}$, is 
named as the $\pi$-noise generator, denoted by $G_{\bm \theta}$. 
The base model is denoted by $h_{\bm \phi}$ where $\bm \phi$ is its learnable parameters. 
The whole procedure is illustrated in Figure \ref{figure_framework}. 

\textcolor{black}{
\textbf{Remark}: 
If the noise $\bm \varepsilon$ is pixel-wise, then there usually exist trivial solutions of $G_{\bm \theta}$, 
For example, $G_{\bm \theta}$ can be identity mapping, \textit{i.e.}, $(\bm x, y) \mapsto \bm x$. 
Then $I(\mathcal{T}, \mathcal{E})$ will be large enough. 
To avoid it, we usually restrict the solution space of $G_{\bm \theta}$, such as only learning variance when $\bm \varepsilon$ is sampled from Gaussian distributions. 
More details can be found in the end of Section \ref{section_gaussian}. 
}

\subsection{Define Task Entropy over Distribution \texorpdfstring{$\mathcal{D}_{\mathcal{X}}$}{} }
To begin with, we first define task entropy over the data distribution, 
which is represented by $\mathcal{D}_{\mathcal{X}}$, 
\begin{equation}
  \begin{aligned}
  H(\mathcal T) 
  & = \mathbb{E}_{\bm x \sim \mathcal{D}_{\mathcal{X}}} H(p(y | \bm x)) 
  = \int - p(\bm x) p(y | \bm x) \log p(y | \bm x) d \bm x d y . 
  \end{aligned}
\end{equation}
With the formal definition of $H(\mathcal{T})$, the mutual information over $\mathcal{D}_{\mathcal{X}}$
can be defined as 
\begin{align}
  I(\mathcal T, \mathcal{E}) & = \mathbb{E}_{\bm x \sim \mathcal{D}_{\mathcal{X}}} \int p(y, \bm \varepsilon | \bm x) \log \frac{p(y, \bm \varepsilon | \bm x)}{p(y | \bm x) \cdot p(\bm \varepsilon | \bm x)} dy d\bm \varepsilon \notag\\ 
  & = \int p(\bm x) p(y, \bm \varepsilon | \bm x) \log \frac{p(y, \bm \varepsilon | \bm x)}{p( y| \bm x) \cdot p(\bm \varepsilon | \bm x)} dy d\bm \varepsilon d\bm x \notag \\
  & = \int p(\bm x) p(y, \bm \varepsilon | \bm x) \log \frac{p(y | \bm x, \bm \varepsilon)}{p(y | \bm x)} dy d\bm \varepsilon d\bm x . 
\end{align}
It should be clarified that we do not define the task entropy on the given 
dataset $\{\bm x_i\}_{i=1}^n$ as $\sum_{i=1}^n H(p(y | \bm x_i)) = -\sum_{i=1}^n \int p(y | \bm x_i) \log p(y | \bm x_i) dy$, 
which is easily extended from Eq. (\ref{eq_initial_definition_of_task_entropy}). 
The advantages to define $H(\mathcal{T})$ on $\mathcal{D}_{\mathcal{X}}$ are mainly from two aspects. 
On the one hand, the definition on a data distribution is a more reasonable setting. 
On the other hand, the expectation can lead to a brief mathematical formulation 
after applying the Monte Carlo method, which will be shown in Eq. (\ref{eq_L_approximate}). 
By reformulating $I(\mathcal{T}, \mathcal{E})$ as  
\begin{align}
  I(\mathcal T, \mathcal{E}) 
  = & \int p(\bm x) p(y, \bm \varepsilon | \bm x) \log p(y | \bm x, \bm \varepsilon) dy d\bm \varepsilon d\bm x \notag \\
  & - \int p(y, \bm x) \log p(y | \bm x) dy d\bm x  \\
  = & \int p(\bm x) p(y, \bm \varepsilon | \bm x) \log p(y | \bm x, \bm \varepsilon) dy d\bm \varepsilon d\bm x + H(\mathcal{T}) , \notag  
\end{align}
we can obtain the formal definition of the conditional task entropy given noise as 
$H(\mathcal{T} | \mathcal{E}) = - \int p(\bm x) p(y, \bm \varepsilon | \bm x) \log p(y | \bm x, \bm \varepsilon) dy d\bm \varepsilon d\bm x$.

\subsection{Variational Approximation}
To overcome \textit{C1} proposed at the end of Section \ref{section_intro}, we turn to maximize a variational lower bound 
of the mutual information by expanding 
$I(\mathcal{T}, \mathcal{E})$ as 
\begin{align}
  I(\mathcal T, \mathcal{E}) 
  \geq &  \int  p(\bm x) p(y, \bm \varepsilon | \bm x) \log q(y | \bm x, \bm \varepsilon) dy d\bm \varepsilon d\bm x \notag \\
  & - \int p(\bm x) p(y | \bm x) \log p(y | \bm x) dy d\bm x \label{ineq_variational_transformation} \\
  = & \int  p(y, \bm x) p(\bm \varepsilon | y, \bm x) \log q(y | \bm x, \bm \varepsilon) dy d\bm \varepsilon d\bm x + H(\mathcal T) , \notag
\end{align}
where $q(y | \bm x, \bm \varepsilon)$ is the tractable variational approximation of $p(y | \bm x, \bm \varepsilon)$. 
Inequality (\ref{ineq_variational_transformation}) is derived from the non-negative property of Kullback-Leibler divergence (KL-divergence) \cite{KL-Divergence}, 
\begin{equation}
KL(p||q) \geq 0 \Leftrightarrow \int p(x) \log p(x) dx \geq \int p(x) \log q(x) dx .
\end{equation}
Since $\bm \varepsilon$ is the learnable variable and $H(\mathcal{T})$ is a constant term during optimization, 
the original problem is equivalent to maximize  
\begin{equation}
  \begin{aligned}
  \mathcal L 
  & = \int  p(y, \bm x) p(\bm \varepsilon | y, \bm x) \log q(y|\bm x, \bm \varepsilon) dy d\bm \varepsilon d\bm x \\
  & = \mathbb{E}_{\bm x, y \sim p(\bm x, y)} \int p(\bm \varepsilon | y, \bm x) \log q(y | \bm x, \bm \varepsilon) d \bm \varepsilon .
  \end{aligned}
\end{equation}
Before the further discussion of $\mathcal L$, it should be emphasized that 
the conditional independence assumption $y \rightarrow \bm x \rightarrow \bm \varepsilon$, 
which is usually applied in related machine learning literatures, 
does not hold in this paper. 
From the mathematical perspective, if $y \rightarrow \bm x \rightarrow \bm \varepsilon$ is a Markov chain, then $p(y | \bm x, \bm \varepsilon) = p(y|\bm x)$, 
which is contradictory with the $\pi$-noise. 
The underlying graphical model is shown in Figure \ref{figure_framework}.

We can form the Monte Carlo estimation of the expectation to avoid the direct 
computation of the integral in $\mathcal{L}$, which is formulated as 
\begin{equation} \label{eq_L_approximate}
  \begin{aligned} 
    \mathcal L 
    & \approx \frac{1}{n} \sum_{i=1}^n \int p(\bm \varepsilon | y_i, \bm x_i) \log q(y_i | \bm x_i, \bm \varepsilon) d\bm \varepsilon \\
    & = \frac{1}{n} \sum _{i=1}^n \mathbb E_{\bm \varepsilon \sim p(\bm \varepsilon | y_i, \bm x_i)} \log q(y_i | \bm x_i, \bm \varepsilon) . 
  \end{aligned}
\end{equation}
Surprisingly, the challenge \textit{C2} is also partially solved by 
avoiding sampling diverse $y$ for $\bm x$. 
Although the above sampling may be biased, it is shown that this approximate method has 
achieved remarkable results from experiments. 

The expectation can be also estimated by the Monte Carlo method to efficiently 
compute the integral of the continuous variable $\bm \varepsilon$. 
To ensure the backpropagation of gradient, we apply the well-known reparameterization trick \cite{VAE} 
and reformulate $p(\bm \varepsilon | y_i, \bm x_i) d \bm \varepsilon = p(\bm \epsilon) d\bm \epsilon$ 
where $\bm \varepsilon = \hat{g}_{\bm \theta}(\bm x_i, y_i, \bm \epsilon)$ is a learnable function
of a supervised sample $(\bm x_i, y_i)$ and a standard multivariate Gaussian random variable $\bm \epsilon \sim p(\bm \epsilon) = \mathcal{N}(0, \bm I)$. 
Accordingly, the objective of variational $\pi$-noise (\textit{VPN}) is 
\begin{equation}
  \begin{aligned}
  \max_{\bm \theta} & \frac{1}{n} \sum _{i=1}^n \int p(\bm \epsilon) \log q(y_i | \bm x_i, \hat{g}_{\bm \theta}(\bm x_i, y_i, \bm \epsilon)) d \bm \epsilon \\
  & = \frac{1}{n} \sum_{i=1}^n \mathbb{E}_{\bm \epsilon \sim p(\bm \epsilon)} \log q(y_i | \bm x_i, \hat{g}_{\bm \theta}(\bm x_i, y_i, \bm \epsilon)) . 
  \end{aligned}
\end{equation}
The above objective can be further written as the Monte Carlo approximation form 
and the loss of VPN is formulated as 
\begin{equation}
  \min_{\bm \theta} \mathcal{L}_{\rm VPN} = -\frac{1}{n \cdot m}\sum_{i=1}^n \sum_{j=1}^{m} \log q(y_i | \bm x_i, \hat{g}_{\bm \theta}(\bm x_i, y_i, \bm \epsilon_{i,j})) , 
\end{equation}
where $\bm \epsilon_{i,j} \sim p(\bm \epsilon)$.
The $\pi$-noise generator $G_{\bm \theta}$ can be accordingly trained by $\mathcal{L}_{\rm VPN}$ if the base model $h_{\bm \phi}$ is frozen when training $G_{\bm \theta}$. 
If the base model is also trained (or fine-tuned) with the generator, 
the final loss consists of two components, 
\begin{equation}
\mathcal{L} = \mathcal{L}_{\rm base} + \gamma \mathcal{L}_{\rm VPN} , 
\end{equation}
where $\gamma$ is a trade-off coefficient and 
$\mathcal{L}_{\rm base}$ represents the original training loss of the base model $h_{\bm \phi}$. 

Finally,  we formally present the relation between $\mathcal{D}_{\bm \theta}$ and the function $\hat{g}_{\bm \theta}$, 
to provide the formulation of $\mathcal{D}_{\bm \theta}$ defined at the beginning of Section \ref{section_method}, as follows
\begin{equation} \label{eq_pe}
  \mathcal{D}_\theta (\bm x_i, y_i) = p(\bm \varepsilon | \bm x_i, y_i) = \int_E^{(L)} p (\bm \epsilon) d\bm \epsilon, 
\end{equation}
where $E = \{\bm \epsilon | \hat{g}_{\bm \theta}(\bm x_i, y_i, \bm \epsilon) = \bm \varepsilon\}$ is a Lebesgue measurable set.

\subsection{Positive-incentive Gaussian $\pi$-Noise} \label{section_gaussian}
To be specific, we elaborate on how to set $p(\bm \varepsilon| y_i, \bm x_i)$ and $q(y_i | \bm x_i, \hat{g}_{\bm \theta}(\bm x_i, y_i, \bm \epsilon_{i,j}))$ used in practice. 
Note that $q(y_i | \bm x_i, \bm \varepsilon)$, the approximation of $p(y_i | \bm x_i, \bm \varepsilon)$, 
should be tractable. 
A simple but effective scheme is to model it by the probability output by the base model, 
\begin{equation} \label{eq_q}
q(y_i | \bm x_i, \bm \varepsilon) = h_{\bm \phi}(\bm x_i, \bm \varepsilon) = {\rm softmax}(\hat{h}_{\bm \phi}(\bm x_i, \bm \varepsilon)) , 
\end{equation}
where $\hat{h}_{\bm \phi}(\cdot)$ represents the neural network without the final decision layer. 
Clearly, most deep models for classification (\textit{e.g.}, ResNet \cite{ResNet}, ViT \cite{ViT}) can be valid $h_{\bm \phi}$. 
With some simple additional modules, some contrastive models (\textit{e.g.}, CLIP \cite{CLIP}) are also compatible with the proposed VPN generator. 
Although there are plenty of techniques (\textit{e.g.}, Siamese network and concatenation) 
to simultaneously process both $\bm x$ and $\bm \varepsilon$ and feed them to the base model, 
a simpler scheme, $\bm x + \bm \varepsilon$, is used in this paper. 
It is the most advantage to avoid refining the architecture of base models 
so that the VPN generator can be easily applied to any network. 

\begin{algorithm}[t]
  \caption{Train base model and the VPN generator to generate Gaussian $\pi$-noise.}
  \label{alg_procedure}
  \begin{algorithmic}
      \REQUIRE Training set $\{(\bm x_i, y_i)\}_{i=1}^n$, batch size $b$, noise size $m$. 
      \ENSURE Generator $G_{\bm \theta}$ and base model $h_{\bm \phi}$. 
      \STATE Initialize the base model $h_{\bm \phi}$ and VPN generator $G_{\bm \theta}$. 
      \FOR {sampled mini-batch $\{(\bm x_k, y_k)\}_{k=1}^b$}{
          \FOR{each $(\bm x_k, y_k)$}{
              \STATE $\bm \Sigma_k = f_{\bm \theta}(\bm x_k, y_k)$ where $\bm \Sigma \in \mathbb{D}^d$. 
              \STATE Sample $m$ noise $\{\bm \epsilon_{k,j}\}_{j=1}^m$ for $\bm x_k$ from $\mathcal{N}(0, \bm I)$. 
              \STATE Reparameterization: $\forall j, \bm \varepsilon_{k,j} = \bm \epsilon_{k,j} \odot {\rm diag}(\bm \Sigma_k)$. 
          }\ENDFOR
          \STATE Obtain $\{(\bm x_k, y_k, \bm \varepsilon_{k,j})\}_{k,j}$. \textcolor{codeAnnotation}{\# Totally $m \times b$ triplets}
          \STATE Update $\bm \theta$ and $\bm \phi$ by minimizing $\mathcal{L} = \mathcal{L}_{\rm base} + \gamma \mathcal{L}_{\rm VPN}$. 
      }\ENDFOR
  \end{algorithmic}
\end{algorithm}
\begin{algorithm}[t]
  \caption{Test a new sample with the trained VPN generator.}
  \label{alg_procedure_test}
  \begin{algorithmic}
      \REQUIRE The trained $h_{\bm \phi}$ and $G_{\bm \theta}$, a test sample $\bm x$. 
      \ENSURE Prediction $y$ of $\bm x$. 
      \FOR{each $Y \in \mathcal{Y}$}{
      \STATE \textcolor{codeAnnotation}{\# Module $G_{\bm \theta} = {\rm Sample} \circ f_{\bm \theta}$ and $\bm \varepsilon_Y  = G_{\bm \theta}(\bm x, Y)$.}
      \STATE Compute $\bm \Sigma_Y = f_{\bm \theta}(\bm x, Y)$. 
      \STATE Sample $\bm \varepsilon_Y \sim \mathcal{N}(0, \bm \Sigma_Y)$. 
      }\ENDFOR
      \STATE $y \leftarrow \arg \max_{Y} q(y=Y | \bm x, \bm \varepsilon_Y)$.  \textcolor{codeAnnotation}{~ \# Using $h_{\bm \phi} (\bm x, \bm \varepsilon_Y)$ referring to Eq. (\ref{eq_q}). }
  \end{algorithmic}
\end{algorithm}

\begin{table*}[t]
  \centering 
  \renewcommand\arraystretch{1.05}
  \small
  \setlength{\tabcolsep}{2.5mm}
  \caption{Test accuracy: All base models $h_{\bm \phi}$ are \textit{jointly} trained with VPN generators $G_{\bm \theta}$. We use ``Res''
  to denote ResNet for simplicity. Note that VPN outperforms all strong augmentations other than Fashion-MNIST with ResNet18 and ResNet34 as base models. 
  It may be caused by the clean background of Fashion-MNIST while the background is real and complicated in all the other datasts.}
  \label{table_accuracy_learnable}
  \begin{tabular}{l c c c c  c c c c c c c}
      \hline
      
      \hline
      \multirow{2}{*}{\diagbox{$G_{\bm \theta}$}{$h_{\bm \phi}$}} & \multicolumn{4}{c}{Fashion-MNIST} & \multicolumn{5}{|c}{CIFAR-10} \\
      \cline{2-10}
      & SR & DNN3 & Res18 & Res34  & \multicolumn{1}{|c}{SR} & DNN3 & Res18 & Res34 & ViT \\
      \hline
      Baseline & 74.97 & 81.57 & 93.60 & 93.84 & 39.08 & 46.92 & 94.77 & 95.00 & 79.91 \\
      \hline
      \textcolor{black}{Random Rotation} & 68.12 & 77.92 & 94.62 & 94.53 & 31.37 & 32.62 & 89.87 & 92.48 & 70.07 \\
      \textcolor{black}{Random Translation} & 67.65 & 78.12 &\textbf{95.33} & \textbf{95.04} & 33.00 & 34.97 & 94.91 & 85.20 & 78.55 \\
      \textcolor{black}{Color Jitter} & 74.98 & 78.52 & 93.79 & 93.84 & 34.40 & 36.47 & 95.04 & 95.23 & 81.01 \\
      \textcolor{black}{Gaussian Blur} & 74.14 & 81.95 & 93.58 & 93.31 & 35.51 & 37.96 & 94.35 & 94.71 & 77.49 \\
      Random Noise & 71.82 & 79.45 & 93.77 & 93.39 & 39.06 & 48.15 & 95.09 & 95.20 & 64.38 \\
      \rowcolor{rowBackgroundColor}DNN3 & 76.34 & 81.79 & 93.84 & 94.23 & 39.36 & 48.76 & \textbf{95.35} & 95.26 & 81.07 \\
      \rowcolor{rowBackgroundColor}DNN3 (No Noise) & 76.34 & 81.83 & 93.84 & 94.23 & \textbf{39.47} & 49.07 & \textbf{95.35} & 95.26 & 81.06 \\
      \rowcolor{rowBackgroundColor}ResNet18 & 67.63 & \textbf{82.19} & 93.85 & 93.89 & 39.34 & \textbf{50.15} & 95.12 & \textbf{95.56} & \textbf{82.26} \\
      \rowcolor{rowBackgroundColor}ResNet18 (No Noise) & \textbf{76.48} & \textbf{82.19} & 93.85 & 93.89 & 39.39 & 50.11 & 95.12 & \textbf{95.56} & \textbf{82.26} \\

      \hline

      \hline
      
  \end{tabular}
\end{table*}

\begin{table}[t]
  \centering 
  \small
  \renewcommand\arraystretch{1.05}
  \setlength{\tabcolsep}{2.2mm}
  \caption{Test accuracy on Tiny ImageNet: Base models and generators are simultaneously trained.}
  \label{table_accuracy_tiny_imagenet}
  \begin{tabular}{l c c c c c c c c c}
      \hline
      
      \hline
      \diagbox{$G_{\bm \theta}$}{$h_{\bm \phi}$} & Res18 & Res34 & Res50 & ViT \\
      \hline
      Baseline & 28.20 & 29.73 & 34.03 & 58.07 \\
      \hline
      \textcolor{black}{Random Rotation} & 34.01 & 33.45 & 32.04 & 57.96 \\
      \textcolor{black}{Random Translation} & 37.87 & 36.55 & 38.02 & 60.92 \\
      \textcolor{black}{Color Jitter} & 28.79& 28.28 & 34.01 & 58.87 \\
      \textcolor{black}{Gaussian Blur} & 28.61 & 28.41 & 35.17 & 57.61 \\
      Random Noise & 32.18 & 32.30 & 35.21 & 40.62 \\
      \rowcolor{rowBackgroundColor}DNN3 & 33.81 & 33.65 & 35.92 & 56.80 \\
      \rowcolor{rowBackgroundColor}DNN3 (No Noise) & \textbf{34.13} & 33.88 & \textbf{36.62} & 56.70 \\
      \rowcolor{rowBackgroundColor}Res18 & 33.05 & \textbf{34.20} & 34.64 & \textbf{61.49} \\
      \rowcolor{rowBackgroundColor}Res18 (No Noise) & 32.42 & \textbf{34.20} & 34.64 & \textbf{61.49} \\
      \hline

      \hline
      
  \end{tabular}
  \vspace{-3mm}
\end{table}

$p(\bm \varepsilon | y_i, \bm x_i)$ is another crucial distribution. 
We assume that $p(\varepsilon | y_i, \bm x_i)$ is an uncorrelated multivariate Gaussian distribution, 
i.e., 
\begin{equation}
p(\bm \varepsilon | y_i, \bm x_i) = \mathcal N(0, \bm \Sigma_i), ~~ s.t. ~ \bm \Sigma_i \in \mathbb{D}^d , 
\end{equation}
where $\mathbb{D}^d$ represents the set of $d \times d$ diagonal matrices. 
The major benefit brought by the assumption of uncorrelation is 
the great reduction of the amount of parameters and burden of computation, 
i.e., from $\mathcal{O}(d^2)$ to $\mathcal{O}(d)$. 
One may argue that $\bm \varepsilon$ \textit{acts more like a data augmentation}. 
Note that We constrain the mean vector as $0$, 
which is another crucial assumption in our paper. 
It effectively \textbf{prevents the generator from falling into a trivial solution}, 
$\bm \mu_i = \bm x_i$ and $\bm \Sigma_i = 0$. 
The constraint ensures that $\bm \varepsilon$ is a random disturbance. 
\textcolor{black}{To control the influence of noise, we can apply further limitations to it. 
For example, if $\pi$-noise obeys the multivariate Gaussian distribution, 
the simple restriction $C_1 \leq \|\bm \Sigma\| \leq C_2$ can efficiently constrain 
the intensity of $\bm \varepsilon$.}

$\bm \Sigma$ is the output of a neural network denoted by $f_{\bm \theta}$, 
\textit{i.e.}, $\bm \Sigma = f_{\bm \theta}(\bm x, y)$. 
In summary, the generator module is composed of sampling and distribution parameter inference, 
\textit{i.e.}, $G_{\bm \theta} = {\rm Sample} \circ f_{\bm \theta}$. 
It should be emphasized that $f_{\bm \theta}$ is a special module different from the existing 
networks, since it takes both $\bm x$ and $y$ as the input 
while the conventional neural networks only require $\bm x$ as the input. 
We try to keep the original architectures of existing well-known networks 
for the sake of stable performance. 
For simplicity, we first convert $(\bm x, y)$ to $\hat{\bm x} = \bm x + \gamma \cdot y$ 
and then feed it to networks. 
$y$ serves as a bias and 
$\gamma$ is a constant coefficient. 

\begin{table}[t]
  \centering 
  \small

  \setlength{\tabcolsep}{3mm}
  \caption{Test accuracy on ILSVRC2012: We load the trained parameters for base models and then fine-tune 
  base models when training generators. Note that the models fine-tuned with ``Color Jitter'' collapse so the results are not reported. }
  \label{table_accuracy_imagenet}
  \begin{tabular}{l c c c c c}
      \hline
      
      \hline
      \diagbox{$G_{\bm \theta}$}{$h_{\bm \phi}$}
       & ResNet18 & ResNet34 & ResNet50  \\
      \hline
      Baseline & 69.76 & 73.30 & 76.12 \\
      \hline
      \textcolor{black}{Random Rotation} & 55.19 & 63.12 & 68.62 \\
      \textcolor{black}{Random Translation} & 59.60 & 66.31 & 71.37 \\
      \textcolor{black}{Gaussian Blur} & 58.53 & 65.53 & 70.67 \\
      Random Noise & 50.67 & 58.79 & 61.48 \\
      \rowcolor{rowBackgroundColor}DNN3 & \textbf{69.87} & \textbf{73.42} & 76.11 \\
      \rowcolor{rowBackgroundColor}DNN3 (No Noise) & \textbf{69.87} & \textbf{73.42} & 76.11 \\
      \rowcolor{rowBackgroundColor}Res18 & 69.81 & \textbf{73.42} & \textbf{76.22} \\
      \rowcolor{rowBackgroundColor}Res18 (No Noise) & 69.81 & \textbf{73.42} & \textbf{76.22} \\
      \hline

      \hline
      
  \end{tabular}
      

  \vspace{2mm}
  \setlength{\tabcolsep}{2mm}
  \caption{Test accuracy: All generators are trained on \textit{fixed} base models. 
    ``Fashion'' and ``CIFAR'' are the abbreviations of Fashion-MNIST and CIFAR-10. }
    \label{table_accuracy_fixed}
    \begin{tabular}{l l c c c c}
        \hline
        
        \hline
         & \multirow{2}{*}{\diagbox{$G_{\bm \theta}$}{$h_{\bm \phi}$}} & \multirow{2}{*}{SR} & \multirow{2}{*}{DNN3} & \multirow{2}{*}{Res18}  & \multirow{2}{*}{Res34} \\
        \\
        \hline
         & Baseline & 74.97 & 81.57 & 93.60 & 93.84 \\
        \rowcolor{rowBackgroundColor}\cellcolor{white}& DNN3 & \textbf{75.08} & \textbf{81.86} & 93.69 & \textbf{95.99} \\
        \rowcolor{rowBackgroundColor}\cellcolor{white}\multirow{-3}{*}{Fashion}& Res18 & 75.06 & 81.74 & \textbf{93.72} & 93.96 \\
        \hline 
        & Baseline & 39.08 & 46.92 & 94.77 & 95.00 \\
        \rowcolor{rowBackgroundColor}\cellcolor{white}& DNN3  & 39.14 & 47.30 & 95.46 & \textbf{95.05} \\
        \rowcolor{rowBackgroundColor}\cellcolor{white}\multirow{-3}{*}{CIFAR} & Res18 & \textbf{39.15} & \textbf{47.35} & \textbf{95.49} & 95.02 \\
        \hline
  
        \hline
        
    \end{tabular}
    
    \vspace{-2mm}
\end{table}

Following \cite{VAE}, we can sample $m$ noises 
for each data point from a standard Gaussian distribution. 
The training algorithm is summarized in Algorithm \ref{alg_procedure}. 
When testing an unseen sample, since the generator requires a label as input, 
we suppose that the sample belongs to some class $Y$ and we can obtain a $\pi$-noise 
drawn from $p(\bm \varepsilon | \bm x, y=Y)$. 
Then the noise is fed to the base model with $\bm x$ 
and $q(y=Y | \bm x, \bm \varepsilon)$ is computed. 
This process is repeated for every class in $|\mathcal{Y}|$ 
and the label with the largest probability will be the predicted class. 
The procedure is shown in Algorithm \ref{alg_procedure_test}.

\section{Experiments} \label{section_experiments}
In this section, we present experimental results to testify:  
(1) \textit{whether the base model benefits from training an additional VPN generator};  
(2) \textit{what kind of noise would be the $\pi$-noise that simplifies the task}; 
(3) \textit{how to build a generator with proper architecture}. 
As discussed in preceding sections, there are two modules involved in the experiments, 
base model and VPN generator, which are represented by $h_{\bm \phi}$ and $G_{\bm \theta}$ respectively.

\begin{figure*}[t]
  \centering
  \includegraphics[width=\linewidth]{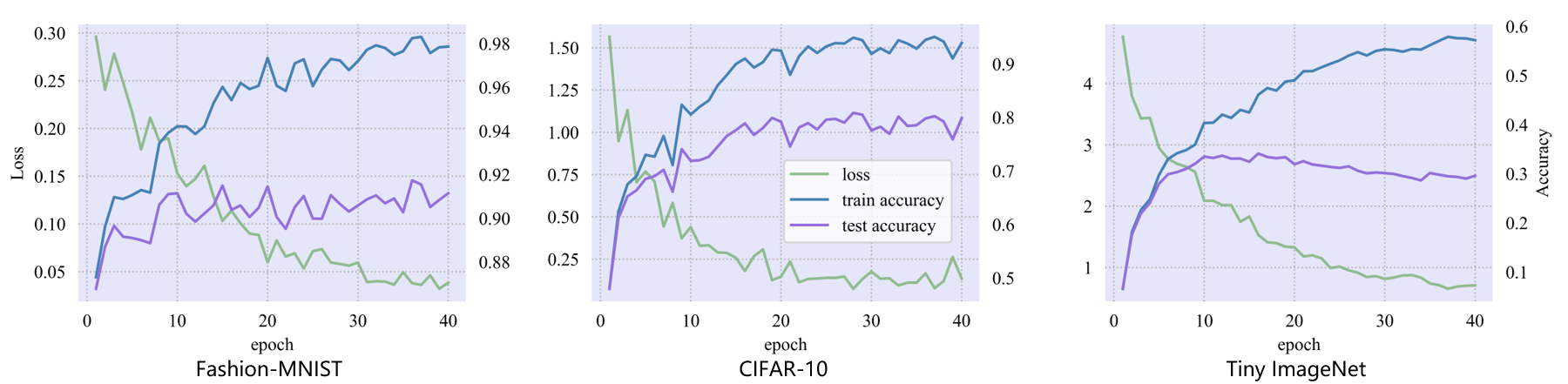}
  \caption{Convergence curve on Fashion-MNIST, CIFAR-10, and Tiny ImageNet, respectively. 
  The base model(ResNet18) and the generator (DNN3) are jointly trained. }
  \label{figure_convergence}
  \vspace{-4mm}
\end{figure*}

\begin{figure}[t]
  \vspace{-2mm}
  \centering
  \hspace{-3mm}
  \subcaptionbox{Fashion-MNIST}{
      \includegraphics[width=0.47\linewidth]{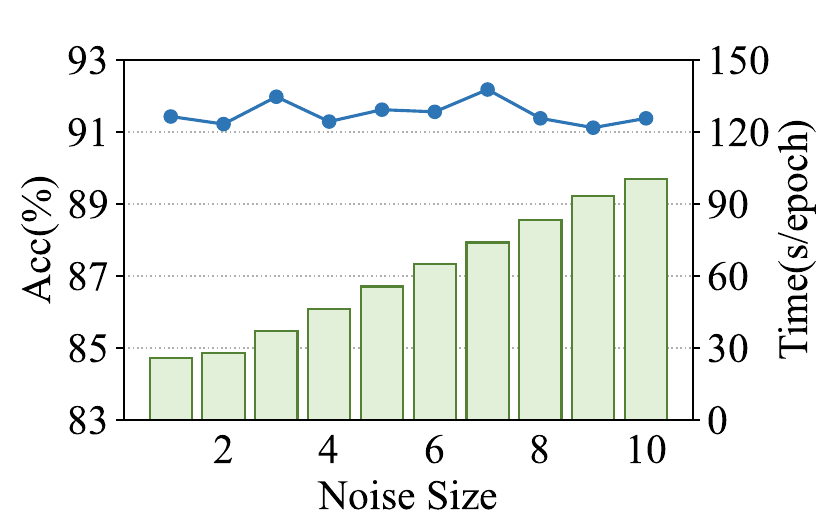}
  }
  \hspace{-3mm}
  \subcaptionbox{CIFAR10}{
      \includegraphics[width=0.47\linewidth]{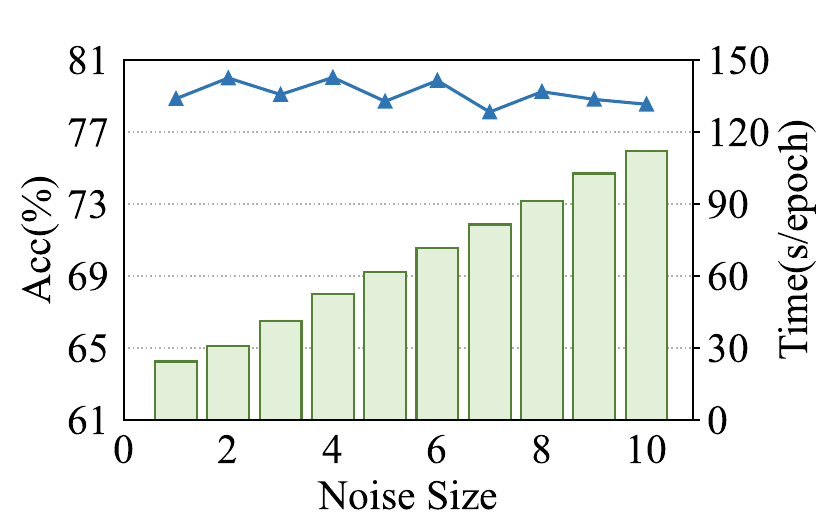}
  }
  \caption{Impact of noise size $m$ on Fashion-MNIST and CIFAR-10. The base model and generator are ResNet18 and DNN3, respectively. }
  \label{figure_m}
  \vspace{-3mm}
\end{figure}

\subsection{Implementation Details}

\textbf{Datasets:} 
Although the proposed VPN framework does not rely on any specific type of data and network, 
we conduct the experiments on vision datasets for the better visualization. 
The idea are validated on Fashion-MNIST \cite{FashionMNIST}, CIFAR-10 \cite{CIFAR10}, 
Tiny ImageNet (a subset of ImageNet Large Scale Visual Recognition Challenge 2012 (ILSVRC2012) \cite{ImageNet}), 
and ILSVRC2012 \cite{ImageNet}. 
Fashion-MNIST and CIFAR-10 contain 70,000 images belonging to 10 classes. 
The split of training-validation-test set is 50,000-10,000-10,000
Tiny ImageNet consists of 110,000 images belonging to 200 categories 
and the ILSVRC2012 contains over 1.2 million images for training and 150,000 images for validation and test. 
There are 500 images for training, 50 images for validation, 
and 50 images for test in each category. 
The image size of Fashion-MNIST, CIFAR-10, and Tiny ImageNet are 
28x28, 32x32, and 64x64, respectively. 
Remark that all input features are firstly scaled to $[0,1]$ before they are fed to neural networks.

\textbf{Settings of base model $h_{\bm \phi}$:} 
On Fashion-MNIST and CIFAR-10, the base models include a linear model, a shallow model, and two deep models. 
Totally 4 different models are employed as the base model in the experiments, including 
the softmax regression (\textit{SR}) \cite{PRML}, 3-layer DNN (\textit{DNN3}), ResNet18 \cite{ResNet}, and ResNet34\cite{ResNet}. 
SR is a linear multi-class classifier extended from the logistic regression 
and is used to show that the VPN generator can be applied to any models  
provided that they can output $p(y | \bm x)$. 
DNN3 serves as a shallow and simple multi-layer perceptron (\textit{MLP}) composed of the form 
$d$-1024-1024-$|\mathcal{Y}|$. 
On Tiny ImageNet and ILSVRC2012, SR and DNN3 are omitted due to the lacking of representative capacity 
and two deeper networks, ResNet50 and ViT \cite{ViT}, are added.

\textbf{Settings of VPN generator $G_{\bm \theta}$:}
The VPN generator intends to generate Gaussian $\pi$-noise which is formally presented 
in Section \ref{section_gaussian}. 
The Gaussian generator $G_{\bm \theta}$ consists of two parts, sampling and parameter learning (denoted by $f_{\theta}$).  
$f_{\bm \theta}$ is also testified with both shallow and deep network, DNN3 and ResNet18. 
Similarly, DNN3 denotes a 3-layer MLP of the form $d$-1024-1024-$d$. 
Remark that the neuron number of the last layer is reduced from $d^2+d$ 
to $d$ owing to the assumption of the zero mean ($\bm \mu=0$) and the uncorrelated variance ($\bm \Sigma \in \mathbb{D}^d$). 

\textbf{Ablation Settings:} 
To testify whether the VPN generator works, we train the base model without any augmentations on three 
datasets under the same settings, which is denoted by \textit{Baseline}. 
We also train the base models with the standard Gaussian noise, 
to validate whether the random Gaussian noise would lead to the same performance improvement. 
Specifically speaking, we randomly select $10\%$ pixels 
and add standard Gaussian noise to them for every image. 
The corresponding results are denoted by \textit{Random}. 
To alleviate the impact of randomness, the corresponding results are 
the averaged result over 5 individual experiments. 
We also report the accuracy of base models, which are trained with generators, 
without feeding noise. The results are denoted by \textit{No Noise}.

\textbf{Hyperparameters:} 
Since how to tune the base model is \textit{not the core purpose} of this paper, 
we simply set the number of epochs as 40 for shallow models 
and 100 for ResNet. 
The learning rate is 0.001. 
To feed both $\bm x$ and $y$ to the generator, 
we process them by $\hat{\bm x} = \bm x + \gamma \cdot y$ 
where $y$ is encoded as the class index starting from 0 and $\gamma = 0.01 \times 1 / |\mathcal{Y}|$. 
On Fashion-MNIST and CIFAR-10, a batch size is set as 256. 
On Tiny ImageNet, the batch size is set as 128 to prevent the OOM exception. 
Another hyperparameter is the noise size $m$, the number of sampling noise during training. 
For the sake of training efficiency, we use $m=1$ on all datasets. 
We run all codes on an NVIDIA GeForce RTX 3090Ti GPU. 
The source code will be released after the formal publication.

      

      

\begin{figure}[t]
  \centering
  \includegraphics[width=\linewidth]{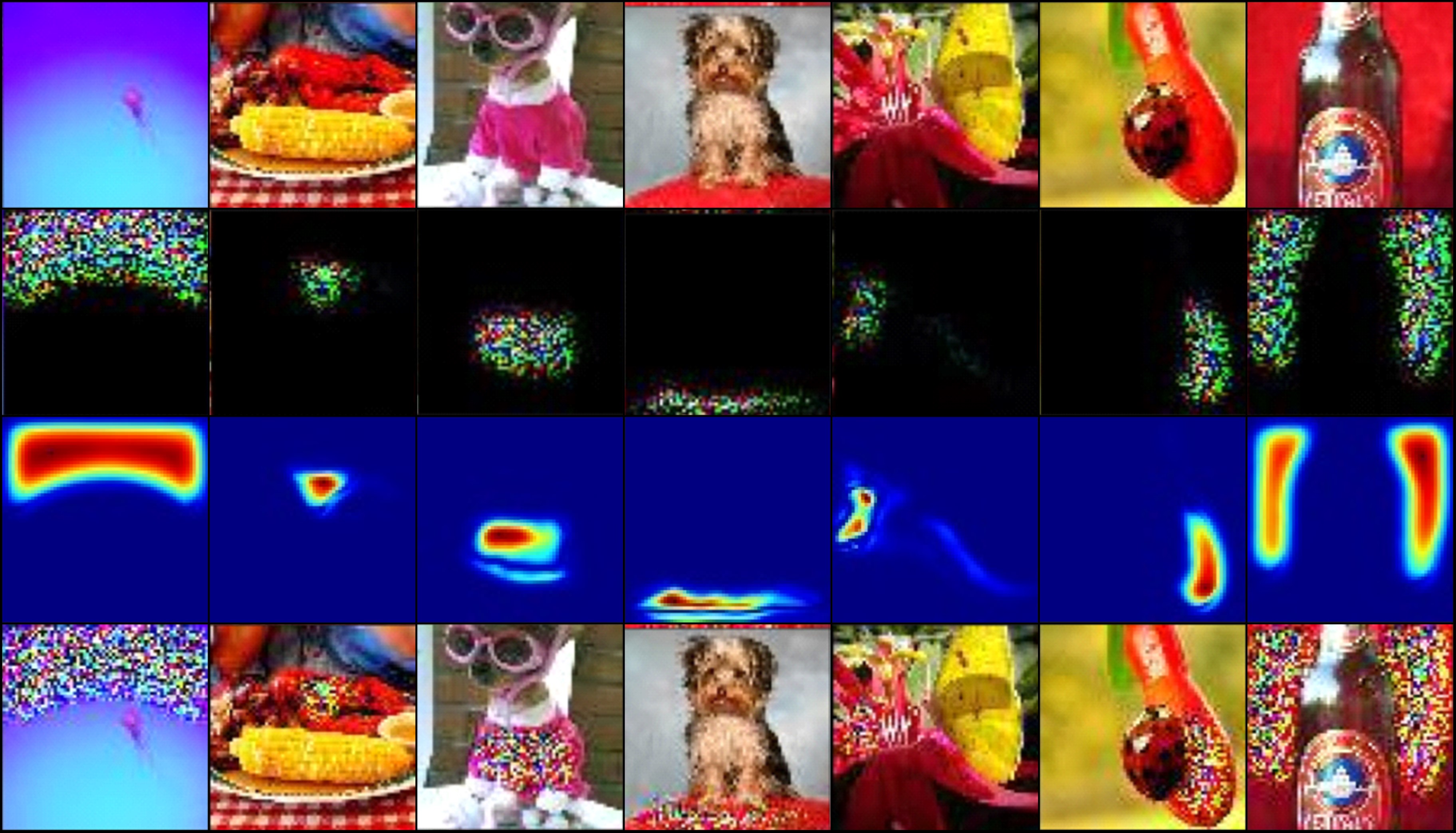}
  \caption{More visualization of generated $\pi$-noise on Tiny ImageNet. 
  The four lines are same as Figure \ref{figure_visualization}.
  }
  \label{figure_visualization_tiny_imagenet}
  \vspace{-4mm}
\end{figure}

\subsection{Performace Analysis}
Overall, the experiments in this paper can be divided into two classes: 
(1) training generators with base models; 
(2) training generators after base models (i.e., frozen base models). 
        
  
        

For the former case, the experimental results on Fashion-MNIST and CIFAR-10 are reported in Table \ref{table_accuracy_learnable}, 
the results on Tiny ImageNet are reported in Table \ref{table_accuracy_tiny_imagenet}, 
and the performance on ILSVRC2012 is reported in Table \ref{table_accuracy_imagenet}. 
Firstly, the generator can promote the base model in most cases. 
Due to the training of generator, the accuracy increases by over $4\%$ on Tiny ImageNet with $h_{\bm \phi}$=ResNet-34 and $G_{\bm \theta}$=ResNet-18.
Secondly, compared with the experiments with random Gaussian noise, 
we confirm that the augmentation by random Gaussian noise is unstable 
and the VPN generator is much more stable. 
It verifies the effectiveness of the idea of learning Gaussian noise. 
Thirdly, DNN3 is more stable than ResNet-18. 
It may be caused by the fact that deeper networks 
easily suffer from over-fitting. 
We also suspect that the information of $\pi$-noise 
may be easier to extract, compared with the visual semantic information, 
so that ResNet-18 may therefore suffer from over-fitting. 
Fourthly, after training base models and generators, 
it seems not necessary to generate $\pi$-noise for unseen samples in the test phase. 
It indirectly validates that the component generated by generators 
is not too intense and does not significantly change the original images. 
\textbf{It is adequate to name it as ``noise''}.

For the latter case that trains generators with frozen base models, 
we show the results of training all generators on 
the trained base models 
in Table \ref{table_accuracy_fixed}. 
Note that when testing a new sample, we sample a noise 
with the help of the trained $G_{\bm \theta}$ so that 
it behaves differently from the trained base model. 
Surprisingly, \textbf{the accuracy improvement is even more stable} 
compared with Table \ref{table_accuracy_learnable}. 
It may be owing to the high sensitivity of training neural networks, 
especially deep models. 
For example, ResNet-50 suffers from a clear degradation on Tiny ImageNet. 
It also implies that the generator may still have the potential value to be further developed.

One may concern that \textit{the inference with the VPN generator will cost too much time 
since the time to generate noise from $p(\bm \varepsilon |\bm x, y)$ is linearly correlated with the class number}. 
From Tables \ref{table_accuracy_learnable}--\ref{table_accuracy_fixed}, 
we conclude that: 
When the base model and VPN generator are jointly trained, 
\textbf{the base model without noise usually achieves similar results} compared with the one with noise 
during the reference phase. 
Even without VPN generator, the base model still benefits from the joint training. 
Therefore we can \textbf{simply drop the VPN generator during the inference phase}, if the computational resources are limited in practice.

\begin{figure}[t]
  \centering
  \includegraphics[width=\linewidth]{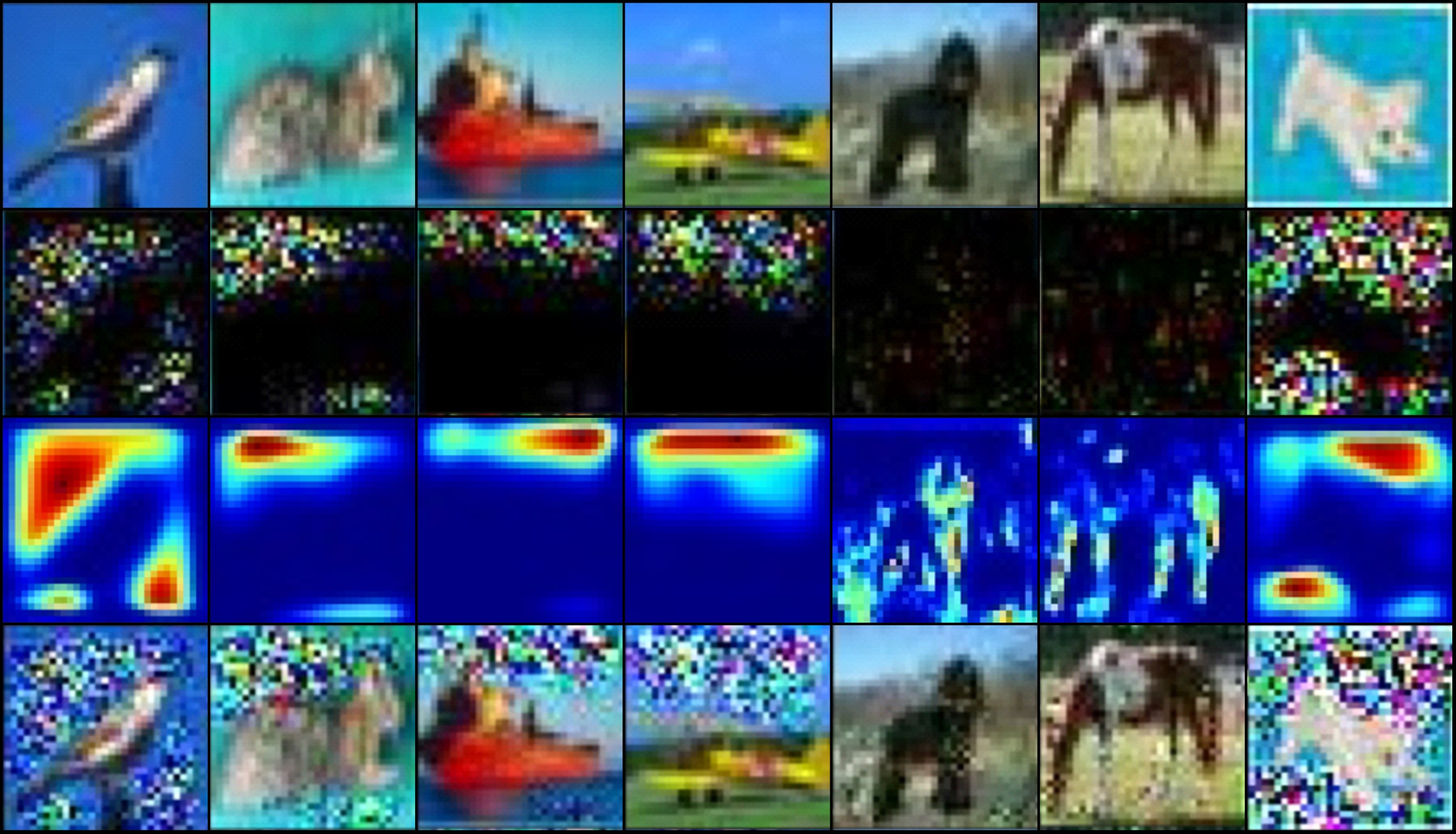}
  \caption{Visualization of generated $\pi$-noise on CIFAR-10.   The four lines are same as Figure \ref{figure_visualization}.}
  \label{cifar-visual}
  \vspace{-4mm}
\end{figure}

\textbf{Convergence:}
We show the convergence curves of loss, training accuracy, and test accuracy in Figure \ref{figure_convergence}. 
From the figure, we find that the training process on Tiny ImageNet is more stable 
compared with Fashion-MNIST and CIFAR-10. 
The oscillation of the curves may be caused by the setting of the noise size $m=1$. 
As elaborated in next paragraph, this setting ensures the fast training of generator without 
apparent performance loss (shown in Figure \ref{figure_m}).

\textbf{Impact of noise size $m$:}
The impact of the number of generated noises, $m$ in Algorithm \ref{alg_procedure}, 
is also investigated and the results are shown in Figure \ref{figure_m}. 
In general, a larger $m$ usually means more stable but slower training. 
From the figure, we surprisingly find that the generator with different $m$ for training achieves 
similar final accuracies, while a larger $m$ leads to faster convergence. 
When the computing resources is limited, we suggest to simply set $m=1$ in practice. 
It should be pointed out that \textit{we use $m=1$ in all other experiments}.

\subsection{Visualization} \label{section_visualization}
To answer the question about ``what kind of noise would be the $\pi$-noise that simplifies the task'', 
we visualize the variance of the learned multivariate Gaussian noise as a heatmap 
and a noise drawn from the distribution in Figure \ref{figure_visualization}. 
In Figure \ref{figure_visualization}, we show two images of Fashion-MNIST and CIFAR-10 
while 5 images from Tiny ImageNet are exhibited. 
From the figure, it is easy to find that 
the learned $\pi$-noise differs on these three datasets. 
More visualization can be found in Figures \ref{figure_visualization_tiny_imagenet} and \ref{cifar-visual} show the visualization images on Tiny ImageNet and CIFAR-10 with the same layout.

Since Fashion-MNIST is a dataset of grayscale images without noisy background, 
it is easy to theoretically speculate that changing background would be helpless 
for decreasing the task entropy defined in Eq. (\ref{eq_initial_definition_of_task_entropy}), 
\textit{i.e.}, $H(\mathcal{T}) = H(\mathcal{T} | \mathcal{E})$. 
Therefore, the $\pi$-noise can only enhance the pixels related to objects. 
The first and second columns in Figure \ref{figure_visualization} strongly support our guess 
and verifies that the proposed VPN indeed tries to minimize the conditional task entropy $H(\mathcal{T} | \mathcal{E})$. 

On CIFAR-10 and ImageNet consisting of colorful images, 
the ideal $\pi$-noise should blur the background irrelevant to the main object 
so that $p(y | \bm x)$ can approach the one-hot distribution 
and the conditional task entropy can decrease. 
From Figure \ref{figure_visualization}, we find that 
the generator aims to disturb the pixels related to background. 
Surprisingly, on more complicated images from Tiny ImageNet, 
the generator precisely distinguishes which pixels are irrelevant. 
For instance, it adds intense noise to the flower in the butterfly image. 
The flower is the ingredient that is most similar to the butterfly. 
From the heatmap images of flagstaff and sea lion, 
the heatmap highly matches the background and the outline of objects is almost detected. 
Even though the base model is not well-trained on Tiny ImageNet, 
the generator has already distinguished the ingredients in images. 
Therefore, the $\pi$-noise may provide a new scheme for segmentation 
and an idea to bridge the classical image recognition and image segmentation.

It should be pointed out that only part of background is detected. 
For example, the seventh image is labeled by flagstaff instead of flag, 
but only the sky is recognized as the irrelevant ingredient. 
It may be caused by the lacking of training data (only 500 images are used for training).

In sum, the visualization verifies that the VPN generator indeed 
helps to enhance the base model and it meets our expectation 
of the task entropy defined in Eq. (\ref{eq_initial_definition_of_task_entropy}).

\section{Future Works} \label{section_limitations}
Owing to the Monte Carlo estimation, the $\pi$-noise generator can be trained 
by sampling without knowing exact $p(y | \bm x)$. 
However, since each data point only has one label in datasets for \textit{single-label} classification, 
we will never get different $y$ for a sample $\bm x$. 
This will lead to a biased estimation of expectations. 
One may argue that multi-label datasets should be used for training models. 
The purpose of defining $p(y | \bm x)$ on single-label classification is 
to measure the task complexity. 
There should be a preference for different classes for $\bm x$. 
However, even on multi-label datasets, $p(y | \bm x)$ is still not provided. 
It is still biased to simply set $p(y | \bm x)$ as a uniform distribution 
since there should be one or two primary labels rather than treating all labels equally.
The rationale is the lacking of suitable datasets that could provide preference ($p(y|\bm x)$)
of different classes but is still designed for classic single-label classification. 
Therefore, it is the principal work to collect a dataset to train $\pi$-noise generators unbiasedly. 

In Section \ref{section_gaussian}, we discuss how to model $q(y | \bm x, \bm \varepsilon)$ 
and $p(\bm \varepsilon | \bm x, y)$, which both require two inputs.
In this paper, they are processed by the simple summation operation to prevent the potential over-fitting risk. 
However, too simple operations may limit the performance and result in bias. 
More sophisticated techniques, especially for $p(\bm \varepsilon | \bm x, y)$, deserve to be studied. 
An ideal scheme is developing a module $\varphi: \mathcal{X} \times \mathcal{Y} \mapsto \mathcal{X}'$ 
where ${\rm dim} ~ \mathcal{X}' = {\rm dim} ~ \mathcal{X}$. 
It is crucial to properly use $y$ without over-fitting as the input of a network to 
generate deep representation. 

It also show the potential to extend $\pi$-noise to other framework such as stochastic systems \cite{Guoliang}, contrastive learning \cite{PiNGDA}, and multimodal learning \cite{PiNI}. 
Specifically, reference \cite{Guoliang} establishes a new and novel theoretical framework for system stability analysis that shows the potential to combine the proposed VPN, and reference \cite{PiNI} shows a promising idea to better model $p(\bm \varepsilon | \bm x, y)$.
              

              

\section{Conclusion}
In this paper, we propose the Variational Positive-incentive Noise (\textit{VPN}), 
an approximate method to learn $\pi$-noise via variational inference, following the framework of $\pi$-noise \cite{Pi-Noise}. 
We validate the efficacy of the idea about generating random noise to enhance the classifier. 
The $\pi$-noise generator always promotes the base model no matter 
whether to train the base model with the generator or not. 
It implies that the VPN generator is a flexible module independent of the existing classifiers. 
From the visualization of generated noise, 
a well-trained $\pi$-noise generator should learn to inject the noise to the irrelevant regions of complicated images 
so that the base model can pay more attention to the right objects, which shows how the noise benefits deep learning models. 

\bibliographystyle{IEEEtran}

\bibliography{citations}

\begin{thebibliography}{10}
\providecommand{\url}[1]{#1}
\csname url@samestyle\endcsname
\providecommand{\newblock}{\relax}
\providecommand{\bibinfo}[2]{#2}
\providecommand{\BIBentrySTDinterwordspacing}{\spaceskip=0pt\relax}
\providecommand{\BIBentryALTinterwordstretchfactor}{4}
\providecommand{\BIBentryALTinterwordspacing}{\spaceskip=\fontdimen2\font plus
\BIBentryALTinterwordstretchfactor\fontdimen3\font minus \fontdimen4\font\relax}
\providecommand{\BIBforeignlanguage}[2]{{%
\expandafter\ifx\csname l@#1\endcsname\relax
\typeout{** WARNING: IEEEtran.bst: No hyphenation pattern has been}%
\typeout{** loaded for the language `#1'. Using the pattern for}%
\typeout{** the default language instead.}%
\else
\language=\csname l@#1\endcsname
\fi
#2}}
\providecommand{\BIBdecl}{\relax}
\BIBdecl

\bibitem{Pi-Noise}
X.~Li, ``Positive-incentive noise,'' \emph{IEEE Transactions on Neural Networks and Learning Systems}, pp. 1--7, 2022.

\bibitem{Dropout}
N.~Srivastava, G.~E. Hinton, A.~Krizhevsky, I.~Sutskever, and R.~Salakhutdinov, ``Dropout: a simple way to prevent neural networks from overfitting,'' \emph{Journal of Machine Learning Research}, vol.~15, no.~1, pp. 1929--1958, 2014.

\bibitem{AdversarialTraining}
C.~Szegedy, W.~Zaremba, I.~Sutskever, J.~Bruna, D.~Erhan, I.~J. Goodfellow, and R.~Fergus, ``Intriguing properties of neural networks,'' in \emph{2nd International Conference on Learning Representations, {ICLR} 2014, Banff, AB, Canada, April 14-16, 2014, Conference Track Proceedings}, 2014.

\bibitem{GAN}
I.~J. Goodfellow, J.~Pouget{-}Abadie, M.~Mirza, B.~Xu, D.~Warde{-}Farley, S.~Ozair, A.~C. Courville, and Y.~Bengio, ``Generative adversarial nets,'' in \emph{Advances in Neural Information Processing Systems 27: Annual Conference on Neural Information Processing Systems 2014, December 8-13 2014, Montreal, Quebec, Canada}, 2014, pp. 2672--2680.

\bibitem{NCE}
M.~Gutmann and A.~Hyv{\"{a}}rinen, ``Noise-contrastive estimation: {A} new estimation principle for unnormalized statistical models,'' in \emph{Proceedings of the Thirteenth International Conference on Artificial Intelligence and Statistics, {AISTATS} 2010, Chia Laguna Resort, Sardinia, Italy, May 13-15, 2010}, vol.~9, 2010, pp. 297--304.

\bibitem{DDPM}
J.~Ho, A.~Jain, and P.~Abbeel, ``Denoising diffusion probabilistic models,'' in \emph{Advances in Neural Information Processing Systems 33: Annual Conference on Neural Information Processing Systems 2020, NeurIPS 2020, December 6-12, 2020, virtual}, 2020.

\bibitem{LabelSmoothing}
R.~M{\"{u}}ller, S.~Kornblith, and G.~E. Hinton, ``When does label smoothing help?'' in \emph{Advances in Neural Information Processing Systems 32: Annual Conference on Neural Information Processing Systems 2019, NeurIPS 2019}, 2019, pp. 4696--4705.

\bibitem{DiffusionNoise}
H.~Qiu, M.~Xia, Y.~Zhang, Y.~He, X.~Wang, Y.~Shan, and Z.~Liu, ``Freenoise: Tuning-free longer video diffusion via noise rescheduling,'' in \emph{The Twelfth International Conference on Learning Representations}, 2024.

\bibitem{ImageTranslationNoise}
C.~Shi, K.~Huang, G.~Lu, H.~Liu, M.~Zhu, N.~Wang, and X.~Gao, ``On the analysis of gan-based image-to-image translation with gaussian noise injection,'' in \emph{The Twelfth International Conference on Learning Representations}, 2024.

\bibitem{PointCloudNoise}
G.~Li, G.~Kang, X.~Wang, Y.~Wei, and Y.~Yang, ``Adversarially masking synthetic to mimic real: Adaptive noise injection for point cloud segmentation adaptation,'' in \emph{Proceedings of the IEEE/CVF Conference on Computer Vision and Pattern Recognition (CVPR)}, June 2023, pp. 20\,464--20\,474.

\bibitem{ImageNet}
O.~Russakovsky, J.~Deng, H.~Su, J.~Krause, S.~Satheesh, S.~Ma, Z.~Huang, A.~Karpathy, A.~Khosla, M.~Bernstein, A.~C. Berg, and L.~Fei-Fei, ``{ImageNet Large Scale Visual Recognition Challenge},'' \emph{International Journal of Computer Vision (IJCV)}, vol. 115, no.~3, pp. 211--252, 2015.

\bibitem{MultiLabel}
G.~Tsoumakas and I.~Katakis, ``Multi-label classification: An overview,'' \emph{Int. J. Data Warehous. Min.}, vol.~3, no.~3, pp. 1--13, 2007.

\bibitem{AutoAugment}
E.~D. Cubuk, B.~Zoph, D.~Man{\'{e}}, V.~Vasudevan, and Q.~V. Le, ``Autoaugment: Learning augmentation strategies from data,'' in \emph{{IEEE} Conference on Computer Vision and Pattern Recognition, {CVPR} 2019}.\hskip 1em plus 0.5em minus 0.4em\relax Computer Vision Foundation / {IEEE}, 2019, pp. 113--123.

\bibitem{CPC}
A.~van~den Oord, Y.~Li, and O.~Vinyals, ``Representation learning with contrastive predictive coding,'' \emph{CoRR}, vol. abs/1807.03748, 2018.

\bibitem{SimCLR}
T.~Chen, S.~Kornblith, M.~Norouzi, and G.~E. Hinton, ``A simple framework for contrastive learning of visual representations,'' in \emph{Proceedings of the 37th International Conference on Machine Learning, {ICML} 2020}, vol. 119, 2020, pp. 1597--1607.

\bibitem{PiNDA}
H.~Zhang, Y.~Xu, S.~Huang, and X.~Li, ``Data augmentation of contrastive learning is estimating positive-incentive noise,'' \emph{arXiv preprint arXiv:2408.09929}, 2024.

\bibitem{VAT}
T.~Miyato, S.-i. Maeda, M.~Koyama, K.~Nakae, and S.~Ishii, ``Distributional smoothing with virtual adversarial training,'' \emph{arXiv preprint arXiv:1507.00677}, 2015.

\bibitem{IMSAT}
W.~Hu, T.~Miyato, S.~Tokui, E.~Matsumoto, and M.~Sugiyama, ``Learning discrete representations via information maximizing self-augmented training,'' in \emph{Proceedings of the 34th International Conference on Machine Learning, {ICML} 2017, Sydney, NSW, Australia, 6-11 August 2017}, vol.~70, 2017, pp. 1558--1567.

\bibitem{FSGM}
I.~J. Goodfellow, J.~Shlens, and C.~Szegedy, ``Explaining and harnessing adversarial examples,'' in \emph{3rd International Conference on Learning Representations, {ICLR} 2015, San Diego, CA, USA, May 7-9, 2015, Conference Track Proceedings}, 2015.

\bibitem{BlackBoxAttack}
J.~Su, D.~V. Vargas, and K.~Sakurai, ``One pixel attack for fooling deep neural networks,'' \emph{{IEEE} Trans. Evol. Comput.}, vol.~23, no.~5, pp. 828--841, 2019.

\bibitem{VAE}
D.~P. Kingma and M.~Welling, ``Auto-encoding variational bayes,'' in \emph{ICLR}, 2014.

\bibitem{VIB}
A.~A. Alemi, I.~Fischer, J.~V. Dillon, and K.~Murphy, ``Deep variational information bottleneck,'' in \emph{5th International Conference on Learning Representations, {ICLR} 2017, Toulon, France, April 24-26, 2017, Conference Track Proceedings}, 2017.

\bibitem{GIB}
T.~Wu, H.~Ren, P.~Li, and J.~Leskovec, ``Graph information bottleneck,'' in \emph{Advances in Neural Information Processing Systems 33: Annual Conference on Neural Information Processing Systems 2020, NeurIPS 2020, December 6-12, 2020, virtual}, 2020.

\bibitem{KL-Divergence}
I.~Csisz{\'a}r, ``I-divergence geometry of probability distributions and minimization problems,'' \emph{The annals of probability}, pp. 146--158, 1975.

\bibitem{ResNet}
K.~He, X.~Zhang, S.~Ren, and J.~Sun, ``Deep residual learning for image recognition,'' in \emph{2016 {IEEE} Conference on Computer Vision and Pattern Recognition, {CVPR} 2016, Las Vegas, NV, USA, June 27-30, 2016}.\hskip 1em plus 0.5em minus 0.4em\relax {IEEE} Computer Society, 2016, pp. 770--778.

\bibitem{ViT}
A.~Dosovitskiy, L.~Beyer, A.~Kolesnikov, D.~Weissenborn, X.~Zhai, T.~Unterthiner, M.~Dehghani, M.~Minderer, G.~Heigold, S.~Gelly, J.~Uszkoreit, and N.~Houlsby, ``An image is worth 16x16 words: Transformers for image recognition at scale,'' in \emph{9th International Conference on Learning Representations, {ICLR} 2021, Virtual Event, Austria, May 3-7, 2021}, 2021.

\bibitem{CLIP}
A.~Radford, J.~W. Kim, C.~Hallacy, A.~Ramesh, G.~Goh, S.~Agarwal, G.~Sastry, A.~Askell, P.~Mishkin, J.~Clark, G.~Krueger, and I.~Sutskever, ``Learning transferable visual models from natural language supervision,'' in \emph{Proceedings of the 38th International Conference on Machine Learning, {ICML} 2021, 18-24 July 2021, Virtual Event}, vol. 139.\hskip 1em plus 0.5em minus 0.4em\relax {PMLR}, 2021, pp. 8748--8763.

\bibitem{FashionMNIST}
H.~Xiao, K.~Rasul, and R.~Vollgraf, ``Fashion-mnist: a novel image dataset for benchmarking machine learning algorithms,'' \emph{arXiv preprint arXiv:1708.07747}, 2017.

\bibitem{CIFAR10}
A.~Krizhevsky, G.~Hinton \emph{et~al.}, ``Learning multiple layers of features from tiny images,'' 2009.

\bibitem{PRML}
C.~M. Bishop and N.~M. Nasrabadi, \emph{Pattern recognition and machine learning}.\hskip 1em plus 0.5em minus 0.4em\relax Springer, 2006, vol.~4, no.~4.

\bibitem{Guoliang}
G.~Chen, C.~Fan, J.~Sun, and J.~Xia, ``Mean square exponential stability analysis for it{\^o} stochastic systems with aperiodic sampling and multiple time-delays,'' \emph{IEEE Transactions on Automatic Control}, vol.~67, no.~5, pp. 2473--2480, 2021.

\bibitem{PiNGDA}
S.~Huang, Y.~Xu, H.~Zhang, and X.~Li, ``Learn beneficial noise as graph augmentation,'' \emph{arXiv preprint arXiv:2505.19024}, 2025.

\bibitem{PiNI}
S.~Huang, H.~Zhang, and X.~Li, ``Enhance vision-language alignment with noise,'' in \emph{Proceedings of the AAAI Conference on Artificial Intelligence}, vol.~39, no.~16, 2025, pp. 17\,449--17\,457.

\end{thebibliography}

\end{document}